\documentclass[letterpaper]{article} 
\usepackage{aaai24}  
\usepackage{times}  
\usepackage{helvet}  
\usepackage{courier}  
\usepackage[hyphens]{url}  
\usepackage{graphicx} 
\usepackage{natbib}  
\usepackage{caption} 
\usepackage{algorithm}
\usepackage{algorithmic}

\usepackage{newfloat}
\usepackage{listings}

\DeclareCaptionStyle{ruled}{labelfont=normalfont,labelsep=colon,strut=off} 
\floatstyle{ruled}
\newfloat{listing}{tb}{lst}{}
\floatname{listing}{Listing}
\usepackage{booktabs}
\usepackage{multirow}
\usepackage{makecell}
\usepackage{amsfonts}
\usepackage{amsthm} 
\usepackage{amssymb} 
\usepackage{amsmath} 
\newtheorem{theorem}{Theorem}

\title{Invariant Random Forest: Tree-Based Model Solution for OOD Generalization}
\author{
    Yufan Liao\textsuperscript{\rm 1,\rm 2},
    Qi Wu\textsuperscript{\rm 2},
    Xing Yan\textsuperscript{\rm 1}\thanks{Corresponding author.}
}
\affiliations{
    \textsuperscript{\rm 1}Institute of Statistics and Big Data, Renmin University of China\\
    \textsuperscript{\rm 2}School of Data Science, City University of Hong Kong\\
    liaoyf@ruc.edu.cn, qi.wu@cityu.edu.hk, xingyan@ruc.edu.cn
}

\begin{document}

\maketitle

\begin{abstract}
Out-Of-Distribution (OOD) generalization is an essential topic in machine learning. However, recent research is only focusing on the corresponding methods for neural networks. This paper introduces a novel and effective solution for OOD generalization of decision tree models, named Invariant Decision Tree (IDT). IDT enforces a penalty term with regard to the unstable/varying behavior of a split across different environments during the growth of the tree. Its ensemble version, the Invariant Random Forest (IRF), is constructed. Our proposed method is motivated by a theoretical result under mild conditions, and validated by numerical tests with both synthetic and real datasets. The superior performance compared to non-OOD tree models implies that considering OOD generalization for tree models is absolutely necessary and should be given more attention.
\end{abstract}
\section{Introduction}

Machine learning models achieve great successes in many applications such as image classification, speech recognition, and so on, when training data and testing data are generated from the same distribution. However, when the theme of the task is about \emph{predicting} rather than \emph{recognizing}, it is a big deal that the distribution of data in testing time may have a huge difference compared to the distribution of training data. Model needs to predict well when the testing data comes from unseen distributions. For example, in autopilot tasks, we may encounter unseen signs; when predicting the stock market, huge crashes may happen for new reasons. This is called the Out-Of-Distribution (OOD) generalization problem. 

The performance of existing machine learning models may drop severely in OOD scenarios \cite{shen2021towards,zhou2021domain,geirhos2020shortcut}. To make up for this weakness, many methods have been proposed to improve the OOD generalization ability of machine learning models, such as IRM \cite{arjovsky2019invariant} and REx \cite{krueger2021out}. But almost all the methods are caring about Deep Neural Networks (DNNs). For other types of models in machine learning, there are no solutions proposed so far.

Decision trees are a type of classic machine learning models. As opposed to DNNs, decision trees are using white-box models which are simple to interpret. Thus, decision trees can be applied to the areas where safety and reliability are required or collaborations are needed between human and AI, such as healthcare, credit assessment, etc. Induced by decision trees, ensemble models like Random Forest (RF) \cite{breiman2001random}, Gradient Boosting Decision Tree (GIDT) \cite{friedman2001greedy}, and XGBoost \cite{chen2015xgboost} become popular choices in real applications.

Though tree-based models have high interpretability, they may also suffer from distribution shifts and spurious correlations, just as DNNs may suffer. Figure \ref{motfigure} gives the illustration of an OOD situation tree models may face. With regular splitting criteria, decision trees only consider the average performance but ignore the heterogeneous behaviors across different environments. Unstable behavior means that the performance of a particular split can change or even twist on the testing set, no matter how good the performance is on training data. When we already know that the data distribution may alter during testing time, it is vital to avoid using the traditional tree models for precautions. 

In this paper, we first describe the failure of decision tree models in the case where testing distribution and training distribution are different. Then, we derive an invariant across multiple environments for classification tasks under the setting of stable and unstable features. Based on this invariant, we introduce our proposed model, Invariant Decision Tree (IDT) and the corresponding ensemble version Invariant Random Forest (IRF), by designing an additional penalty term in the splitting criterion for growing trees. 
The penalty term encourages the use of stable features as the splitting variable in the tree growth.
Experiments on both synthetic datasets and real datasets  prove the superiority of our proposed method in OOD generalization.

\section{Related Works}

\paragraph{Invariant Learning}
To achieve OOD generalization of DNNs, invariant learning methods such as IRM \cite{arjovsky2019invariant} and REx \cite{krueger2021out} propose to pursue invariant final layer or invariant loss across data from different sources (e.g., locations, time, etc.). The data from different sources are usually called different environments or domains in this kind of literature. IRM seeks the invariant of the last layer of the networks, and REx requests that the losses in different environments are close. Invariant learning believes that a model with invariant behaviors in all the training environments will have consistency in the prediction under testing environments.
In this paper, we will discover a new kind of invariant for tree models which is different from those of IRM and REx.


\paragraph{Stable Learning}
Stable learning shares some common motivations with causal inference \cite{cui2022stable}. In the literature, stable prediction was achieved via variable balancing \cite{kuang2018stable}, sample re-weighting \cite{shen2020stable_1}, and  feature decorrelation \cite{shen2020stable}.
It was further extended to problems of adversarial \cite{liu2021stable},  sparse \cite{yu2023stable}, model mis-specification \cite{kuang2020stable}, and deep model for images \cite{zhang2021deep}.

\paragraph{Time Robust Tree}
Time Robust Tree (TRT) \cite{moneda2022time} brought the OOD generalization problem to decision trees.  TRT proposed to restrict the minimum sample size for each environment after a split and consider the maximum loss across environments instead of the empirical loss. However, this method is limited to time series data only, and the performance is not outstanding on real datasets.

\section{Motivation: Stable Split and Unstable Split}

\begin{figure*}
	\centering
	\includegraphics[width = .9\textwidth]{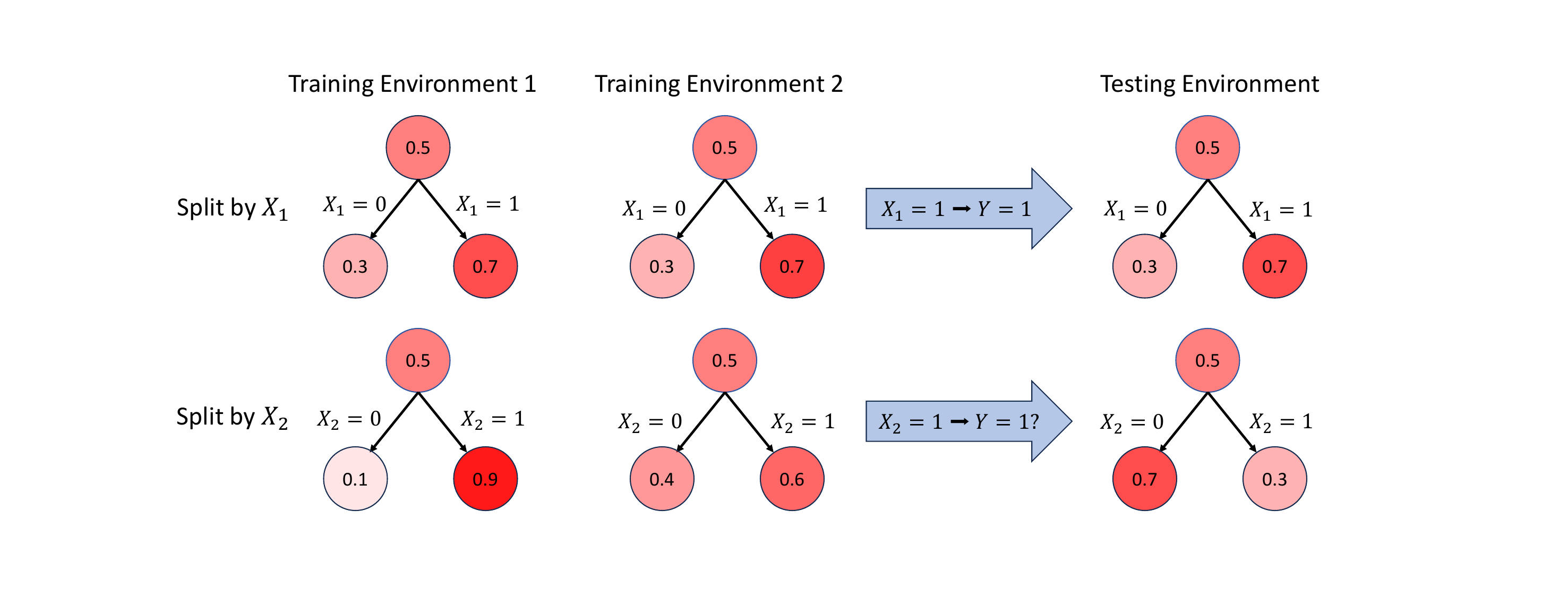}
	\vspace{-2em}
	\caption{The illustration of the example in \eqref{generalMOT}. The number in the circle is $\mathbb{E}[Y]$.
		The prediction rule $X_2 = 1 \rightarrow Y = 1$ has different accuracies under two different training environments, so it is questionable if this rule can still be effective under the testing environment.}
	\label{motfigure}
\end{figure*}

Before formally describing our proposed method, we first raise a toy example to show the motivation of our method. Consider a simple binary classification problem, where the label $Y \in \{0,1\}$, and the 2-dimensional features $X = (X_1,X_2) \in \{ (0,0),(0,1),(1,0),(1,1) \}$ are also binary. Suppose we have two training environments $e=1,2$ with the same sample size, and a testing environment $e=3$. We consider the out-of-distribution case where in different environments, the causal relations between $X_1$ and $Y$ are stable, but the causal relations between $X_2$ and $Y$ vary:
\begin{equation}
\begin{aligned}
    &Y = \text{Bern}(0.5), \\
    &X_1 = |Y - C_1|,\quad C_1 \sim \text{Bern}(0.3),\\
    &X_2 = |Y - C_2|,\quad C_2 \sim \text{Bern}(U_e).
\label{generalMOT}
\end{aligned}
\end{equation}

Here, $\text{Bern}(p)$ refers to the Bernoulli distribution with parameter $p$. Different environments $e$ have different parameters $U_e$. The equations are saying that $X_1$ is equal to label $Y$, but would twist with probability 0.3. $X_2$ is also equal to $Y$, and twist with probability $U_e$. Then, suppose $U_1 = 0.1,~U_2 = 0.4,~U_3 = 0.7$.
When decision tree picks the best split for the training data, it would pick $X_2$ as the splitting variable rather than $X_1$. This is because, in the mixed data of environment 1 and environment 2, $X_2$ equals to label $Y$ at 75\% of the time (for $X_1$, it is 70\%), showing a stronger relation and predictability to label $Y$. 

However, given the environmental information we know, to improve the generalization performance of the model, it is better to use $X_1$ instead of $X_2$ as the predicting variable. The given data-generating process has implied that using $X_1$ as the splitting variable is more secure and can ensure good performance at testing time.  If we normally learn from the training data and derive $X_2 = 1 \rightarrow Y = 1$, then we can only achieve 30\% accuracy on the testing set. But if we derive $X_1 = 1 \rightarrow Y = 1$, it is 70\%. Figure \ref{motfigure} is an illustration of this toy example.

To solve this problem, one needs to avoid the split that behaves differently across different environments. As in the above example, when the split has 90\% accuracy in one environment, and only 60\% in another environment. With such a huge difference, it is questionable whether this split can be effective on testing data. However, how can we rule out these unstable splits systematically and quantitatively? In the next, we will consider a more general data setting, and develop an invariant that can help us discriminate the stable and unstable splits.

\subsection{The Theorem of An Invariant}

Now we consider a general binary classification problem. Similar to the assumption in \citet{rosenfeld2020risks}, we assume that in the data-generating process, the label is drawn first, and features are drawn according to the label. For each environment $e$,
\begin{equation}
    Y = \left\{
    \begin{aligned}
    &    1, \quad \text{w.p.} ~~ \eta^e, \\
    &    0, \quad\text{otherwise}. \\
    \end{aligned}
    \right .
    \label{risk1}
\end{equation}
Stable variables $S = (S_1,S_2,...,S_r)$ and environmental variables $Z = (Z_1,Z_2,...,Z_t)$ are generated according to the label $Y$:
\begin{equation}
\begin{split}
        S \sim P_0(s_1,s_2,...,s_r),~ Z \sim Q_0^e(z_1,z_2,...,z_t),~ \text{if } Y = 0, \\
        S \sim P_1(s_1,s_2,...,s_r),~ Z \sim Q_1^e(z_1,z_2,...,z_t),~ \text{if } Y = 1.    
\end{split}
\end{equation}
Here, $P_0,Q_0^e,P_1,Q_1^e$ are cumulative distribution functions. Note that with the same label, the distributions of stable variables $S$ are the same across different environments, while environmental variables $Z$ are not. We consider a more general setting compared to \citet{rosenfeld2020risks}, where $S$ and $Z$ can follow any distributions instead of only Gaussian distributions. 

The goal of the model is to take both stable and environmental variables $X = (S,Z)$ as inputs, to predict the label $Y$. Decision trees use one splitting variable at a node to separate the data into two subsets. As the distribution of $Z$ may change in a new testing environment, we would like to avoid using $Z$ as the splitting variables in our tree model. However, we do not know which variable is the stable one in advance. We have to judge the stability of the variable on our own. To achieve that, we first derive an invariant across environments when using the stable variables $S$ as the splitting variables. Then, this invariant can be the judging criterion for the stability of any splitting variable.

\begin{theorem}
    For any subset $D \subset \mathbb{R}^{r+t}$, let $\tilde{P_0}, \tilde{Q_0^e}, \tilde{P_1}, \tilde{Q_1^e} $ be the distribution of $P_0,Q_0^e,P_1,Q_1^e$ restricted on $D$, respectively. For a splitting rule $S_i  \leq c$ on $D$, define the changing rate of positive label as
    \begin{equation}
        \text{CR}_{S_i \leq c}^1 = \frac{\mathbb{P}(Y = 1|S_i \leq c,X \in D)}{\mathbb{P}(Y = 1| X \in D)},
    \end{equation}
    and the changing rate of negative label as
    \begin{equation}
        \text{CR}_{S_i \leq c}^0 = \frac{\mathbb{P}(Y = 0|S_i \leq c,X \in D)}{\mathbb{P}(Y = 0| X \in D)}.
    \end{equation}
    These changing rates can be calculated in any environment.
    Then, using any stable variable as the splitting variable, the ratio between the changing rates of positive label and negative label is invariant across different environments. That is to say, for some $k$, $\text{CR}_{S_i \leq c}^1 / \text{CR}_{S_i \leq c}^0 = k$ stands for every environment $e$.
    \label{crthm}
\end{theorem}

\begin{proof}
    We hide the condition $X\in D$ for simplicity. For any environment $e$, the probability of positive label conditional on $S_i  \leq  c$ is $\mathbb{P}(Y = 1 | S_i \leq  c) =  \frac{\mathbb{P}(Y=1) \mathbb{P}(S_i\leq c|Y=1)}{\mathbb{P}(S_i\leq c)} $. Similarly, $\mathbb{P}(Y=0|S_i  \leq c)  = \frac{\mathbb{P}(Y=0) \mathbb{P}(S_i \leq c|Y=0)}{\mathbb{P}(S_i \leq c)} $. Let the first equation be divided by the second one, we have
\begin{equation}
        \frac{\mathbb{P}(Y=1|S_i \leq c)}{\mathbb{P}(Y=0|S_i \leq c)}  = \frac{\mathbb{P}(Y=1) \mathbb{P}(S_i \leq c|Y=1)}{\mathbb{P}(Y=0) \mathbb{P}(S_i \leq c|Y=0)}.
\end{equation}
    Then,
\begin{equation}
    \begin{split}
        & \frac{\mathbb{P}(Y=1|S_i\leq c)}{\mathbb{P}(Y=1)}  / \frac{\mathbb{P}(Y=0|S_i \leq c)}{\mathbb{P}(Y=0)}   \\
        = & \frac{\mathbb{P}(S_i \leq c|Y=1, X\in D)}{\mathbb{P}(S_i \leq c|Y=0, X\in D)} 
        =  \frac{\mathbb{P}_{S \sim \tilde{P_1}}(S_i \leq c)}{\mathbb{P}_{S \sim \tilde{P_0}}(S_i \leq c)}.
    \end{split}
\end{equation}
    Because all the environments share the same distribution of $S$ over $D$, i.e., $\tilde{P_0}$ and $\tilde{P_1}$, the right-hand side of the last equation is invariant across different environments, as long as the splitting variable is a stable variable. 
\end{proof}

The above theorem suggests that, when we use the stable variable as the splitting variable, the ratio of changing rates is an invariant across different environments. It may no longer be an invariant if the splitting variable is an environmental variable, because $Q^e_0$ and $Q^e_1$ are changing over environments. Therefore, in order to use more stable variables as splitting variables in the tree model, we could enforce a restriction using this invariant during the splitting procedure. 

\section{Invariant Tree and Forest}

A decision tree recursively partitions the feature space such that the samples with the same labels or similar target values are grouped together. Letting the training data set at node $m$ be $Q_m$, with sample size $n_m$, and we represent each candidate split $X_j \leq c$ as $\theta = (j,c)$. $\theta$ splits the data set $Q_m$ into $Q_{m,l}(\theta)$ and $Q_{m,r}(\theta)$ with $n_{m,l}$ and $n_{m,r}$ samples respectively:
\begin{equation}
    \begin{split}
        Q_{m,l}(\theta) = \{ (x,y) \in Q_m | x_j \leq c \}, \\
        Q_{m,r}(\theta) = \{ (x,y) \in Q_m | x_j > c \}. \\
    \end{split}
	\label{eqn:l_r_split}
\end{equation}
An impurity function $H(\cdot)$ can measure the similarity of a group of data. Generally, the quality of a candidate split $\theta$ can be measured by calculating the weighted sum of the impurity functions on both left and right nodes:
\begin{equation}
    G(Q_m, \theta) = \frac{n_{m,l}}{n_m} H(Q_{m,l}) +  \frac{n_{m,r}}{n_m} H(Q_{m,r}).
\end{equation}
To pick out the best split, we select the one that minimizes the impurity objective:
\begin{equation}
    \theta^* = \arg \min_\theta G(Q_m, \theta).
\end{equation}
A grid searching for $\theta$  is generally adopted.

For classification tasks, the popular choices of the impurity function are Gini impurity \cite{gordon1984classification} and Shannon information gain \cite{quinlan2014c4}. We use Gini impurity throughout the paper. 
For regression tasks, Mean Squared Error (MSE) and Mean Absolute Error (MAE) can be used as impurity functions. We use MSE as the impurity function in regression tasks throughout the paper. 
Note that for both types of tasks, the impurity functions measure how concentrated a set of data is. 

\subsection{Invariant Classification Tree}

To consider the case of classification tree, first, we denote the data sets with the environments they come from: $Q_m^e$ is the training data set at node $m$ which comes from environment $e$, where $e=1,\dots,E$ and $E$ is the number of training environments. Then, $\theta = (j,c)$ can split $Q_m^e$ into two subsets $Q_{m,l}^e(\theta)$ and $Q_{m,r}^e(\theta)$, as in \eqref{eqn:l_r_split}. Note that the merged data set $Q_m = \cup_{e=1}^T Q_m^e$ is used to compute the impurity objective $G(Q_m, \theta)$ that needs to be minimized in our tree model.


In order to consider the invariance across environments, we propose to put a restriction on the split, together with the original splitting objective $G(Q_m, \theta)$. In Theorem \ref{crthm}, we have defined the changing rates on a set of distributions. Now, we can compute the changing rates given datasets:
    \begin{align}
	\text{CR}_{X_j \leq c}^1(Q_m^e) = \frac{ | \{(x,y) \in Q_{m,l}^e (\theta)|y = 1\}|/|Q_{m,l}^e(\theta)|}{| \{(x,y) \in Q_{m}^e |y = 1\}|/|Q_{m}^e|}, \\
	\text{CR}_{X_j \leq c}^0(Q_m^e) = \frac{ | \{(x,y) \in Q_{m,l}^e (\theta)|y = 0\}|/|Q_{m,l}^e(\theta)|}{| \{(x,y) \in Q_{m}^e |y = 0\}|/|Q_{m}^e|}.        
	\end{align}
Through Theorem \ref{crthm}, when the splitting variable $X_j$ is a stable one, an invariant across environments should be
\begin{equation}
    I(Q_m^e, \theta) = \text{CR}_{X_j \leq c}^1(Q_m^e) / \text{CR}_{X_j \leq c}^0(Q_m^e).
\end{equation}
Therefore, when deciding the best split at a node, we can select the split which makes $I(Q_m^e, \theta)$ unchanged across different environments:
\begin{equation}
\begin{aligned}
    \theta^* & = \arg \min_\theta G(Q_m, \theta), \\
    \text{s.t.} \quad I(Q_m^e, \theta) & = I(Q_m^f, \theta), \quad\forall e,f = 1,2,...,E. 
    \label{originalIRF}
\end{aligned}
\end{equation}
By adding this restriction, we can ensure the invariance of the model across environments. However, the restriction is too strong to be satisfied in reality. 

Thus, we transfer this hard restriction into a penalty term. The more $I(Q_m^e, \theta)$ varies, the bigger the penalty term should be. In this way, the problem turns out to be solvable:
\begin{equation}
    \theta^* = \arg \min_\theta G(Q_m, \theta) + \lambda L(Q_m^1, ..., Q_m^E, \theta),
    \label{eqn:penalized_impurity}
\end{equation}
where $Q_m^e$ is the training data set from the $e$-th environment, $L$ is the penalty term regarding the invariance among environments, and $\lambda \in \mathbb{R}_+$ is the penalty weight. Notice that if $L \geq 0$, then $\lambda \rightarrow \infty$ will push $L \rightarrow 0$. Here, we define $L$ to be the ratio between the highest and the lowest $I(Q_m^e,\theta)$ minus 1, which is
\begin{equation}
    L(Q_m^1,..., Q_m^E, \theta) = \max \limits_{e,f} I(Q_m^e,\theta) / I(Q_m^f,\theta) - 1.
\end{equation}
As a result, when $\lambda$ becomes larger, the difference among $I(Q_m^e,\theta)$ from different environments should be smaller when minimizing the objective function \eqref{eqn:penalized_impurity}. When $\lambda \rightarrow \infty$, $L$ is pushed to be 0, meaning that all $I(Q_m^e,\theta)$ will be the same, which is exactly the hard restriction  in \eqref{originalIRF}.

In the algorithm, to prevent the case when the denominator is 0, we add a dummy sample to the calculation of changing rates:
    \begin{align}
	\tilde{\text{CR}}_{X_j \leq c}^1(Q_m^e) & \propto \frac{ | \{(x,y) \in Q_{m,l}^e (\theta)|y = 1\}| + 0.5 }{| \{(x,y) \in Q_{m}^e |y = 1\}|+1}, \\
	\tilde{\text{CR}}_{X_j \leq c}^0(Q_m^e) & \propto \frac{ | \{(x,y) \in Q_{m,l}^e (\theta)|y = 0\}|+ 0.5}{| \{(x,y) \in Q_{m}^e |y = 0\}|+1}, \\
	\tilde{I}(Q_m^e, \theta) & = \tilde{\text{CR}}_{X_j \leq c}^1(Q_m^e) / \tilde{\text{CR}}_{X_j \leq c}^0(Q_m^e).   
\end{align}
The penalty term $L$ is computed using $\tilde{I}(Q_m^e, \theta)$ instead.

\subsection{Invariant Regression Tree}

In regression tasks, we cannot define changing rates as the ones in the classification problem. However, we can follow the motivation of changing rates, which is the change of average label after a split. Thereby, we propose to define the changing rate in regression problem as the difference of label mean before and after a split:
\begin{align}
    \text{CR}_{X_j \leq c}(Q) & = \frac{\sum \limits_{(x,y) \in Q_{l} (\theta)} y} {|Q_{l}(\theta)|} -  \frac{\sum \limits_{(x,y) \in Q} y} {|Q|}, \\
    I(Q_m^e,\theta) & = \text{CR}_{X_j\leq c}(Q_m^e),
\end{align}
where $Q$ is the training data set at a node in any environment and $Q_{l} (\theta)$ is the left-node data set after applying the split $\theta$ on $Q$. 

For the penalty term, we calculate it as the variance of changing rates from all training environments:
\begin{equation}
    L(Q_m^1,...,Q_m^E, \theta) = \text{Var}_{e=1,\dots,E}[I(Q_m^e,\theta)].
\end{equation}
The final penalized impurity objective for minimization is again the one in \eqref{eqn:penalized_impurity}.
This idea is in part driven by the changing rate definition in classification problem, and is also in part similar with REx  \cite{krueger2021out}. Differently, REx computes the penalty term using the variance of loss functions from different training environments, while our model uses the variance of changing rates.

\subsection{Invariant Random Forest}

Except the penalty term added in the splitting criterion, the rest of the training process is the same as decision trees. Data on a node is recursively split into two subsets until the maximum tree depth is reached or all the data points on a node have a same label. 
Invariant Random Forest (IRF), the ensemble of many invariant classification/regression trees, follows a similar process as aggregating decision trees to Random Forest. We do the bootstrap sampling for the training data, which is separately done for each of the training environments, and generating the output by voting.

Another key difference from traditional Random Forest is that IRF does not extract a random subset of features in splitting.
The random subset of features is a crucial point in Random Forest, aiming at improving the diversity of trees.
But in IRF, the penalty term will avoid using some part of features that are suspicious for unstable splits. This is another kind of feature selection in splitting, and given the diversity of environments, the random feature selection is unnecessary.

\section{Experiments}

\begin{table*}[t]
\begin{center}
\resizebox{\textwidth}{!}{
\begin{tabular}{ccccccccc} 
\toprule
\makecell[c]{Dataset\\ Dimension} & RF & LR & IRM & REx & XGBoost & IRF ($\lambda = 1$)& IRF ($\lambda = 5$)& IRF ($\lambda = 10$) \\
\midrule
$d=2$ & 48.74 (0.92) & 47.22 (0.80) & 44.90 (2.92) & \textbf{52.72 (1.24)} &  49.50 (1.22) & 50.24 (0.96) & 51.20 (0.65) & 51.06 (0.62) \\
$d=5$ & 47.62 (0.61) & 48.62 (0.24) & 47.54 (0.68) & 53.54 (0.46) & 49.14 (0.41) & 52.24 (0.86) & 55.04 (0.82) & \textbf{55.12 (1.22)}   \\
$d=10$ & 43.26 (0.90) & 47.12 (0.40) & 47.86 (1.89) & 53.24 (0.94) & 46.64 (0.42) & 51.26 (1.33) & 53.06 (0.89) & \textbf{54.94 (0.58)}  \\
$d=20$ & 40.08 (0.79) & 51.78 (0.68) & 47.44 (1.01) & 54.56 (0.89) & 48.28 (0.62) & 52.56 (1.02) & 55.08 (0.75) & \textbf{57.42 (0.47)} \\
\bottomrule
\end{tabular}}
\end{center}
\caption{Accuracy results of the synthetic classification task. We consider four cases when the dataset dimension varies in $\{2,5,10,20\}$. The numbers reported are the averages from 5 trials with different randomness seeds. Standard deviations are in the brackets.}
\label{synclassification}
\end{table*}

\begin{table*}[t]
\small
\begin{center}
\begin{tabular}{ccccccccc} 
\toprule
\makecell[c]{Dataset\\ Dimension} & RF & LR & IRM & REx & XGBoost & IRF ($\lambda = 1$)& IRF ($\lambda = 5$)& IRF ($\lambda = 10$) \\
\midrule
$d=2$ & 1.0 & 2.095 (0.056) & 1.014 (0.027) & 1.076 (0.062) & 1.131 (0.020) & 0.982 (0.014) & 0.914 (0.020) & \textbf{0.796 (0.014)} \\
$d=5$ & 1.0 & 1.779 (0.016) & 0.786 (0.031) & 0.928 (0.018) & 1.028 (0.019) & 1.015 (0.010) & 0.956 (0.019) & \textbf{0.761 (0.022)} \\
$d=10$ & 1.0 & 1.486 (0.030) & 0.864 (0.033) & 0.807 (0.013) & 1.008 (0.042) & 1.022 (0.009) & 0.930 (0.014) & \textbf{0.721 (0.016)} \\
$d=20$ & 1.0 & 1.348 (0.029) & 0.895 (0.022) & 0.818 (0.015) & 1.066 (0.018) & 0.990 (0.009) & 0.874 (0.008) & \textbf{0.684 (0.011)} \\
\bottomrule
\end{tabular}
\end{center}
\caption{MSE results of the synthetic regression task. We consider four cases when the dataset dimension varies in $\{2,5,10,20\}$. The numbers reported are the averages from 5 trials with different randomness seeds. Standard deviations are in the brackets.}
\label{synregression}
\end{table*}

\begin{table*}[t]
	\begin{center}
		\resizebox{\textwidth}{!}{
		\begin{tabular}{ccccccccc} 
			\toprule
			\multicolumn{1}{c}{Dataset} & \multicolumn{2}{c}{RF (IRF with $\lambda = 0$)} & \multicolumn{2}{c}{ IRF ($\lambda = 1$)} & \multicolumn{2}{c}{ IRF ($\lambda = 5$)}& \multicolumn{2}{c}{ IRF ($\lambda = 10$)} \\
			\cmidrule(r){2-3} \cmidrule(r){4-5} \cmidrule(r){6-7} \cmidrule(r){8-9}
			& Stable & \makecell[c]{Environ-\\ mental} & Stable & \makecell[c]{Environ-\\ mental} & Stable & \makecell[c]{Environ-\\ mental} & Stable & \makecell[c]{Environ-\\ mental} \\
			\midrule
			Classification ($d=2$) & 4.25 (0.14) & 4.76 (0.12) & 6.06 (0.46) & 3.42 (0.45) & 6.44 (0.43) & 3.10 (0.42) & 6.54 (0.40) & 3.02 (0.39)\\
			Classification ($d=5$) & 3.84 (0.14) & 4.83 (0.12) & 5.67 (0.30) & 3.61 (0.29) & 6.25 (0.29) & 3.20 (0.29) & 6.38 (0.34) & 3.11 (0.34)\\
			Classification ($d=10$) & 3.22 (0.11) & 5.18 (0.15) & 5.26 (0.30) & 3.83  (0.30) & 5.86 (0.23) & 3.46 (0.23) & 6.12 (0.19) & 3.27 (0.19)\\
			Classification ($d=20$) & 2.84 (0.05) & 5.44 (0.05) & 5.26 (0.23) & 3.64 (0.22) & 5.70 (0.20) & 3.43 (0.19) & 5.90 (0.21) & 3.31 (0.21)\\
			Regression ($d=2$) & 6.54 (0.03) & 4.87 (0.02) & 6.69 (0.18) & 4.65 (0.16) & 8.15 (0.18) & 3.56(0.22) & 9.26 (0.08) & 2.17 (0.11)\\
			Regression ($d=5$) & 7.52 (0.02) & 4.01 (0.02) & 7.43 (0.07) & 3.91 (0.07) & 8.75 (0.09) & 3.00 (0.06) & 9.72 (0.06) & 1.75 (0.06)\\
			Regression ($d=10$) & 7.95 (0.02) & 3.65 (0.01) & 7.88 (0.05) & 3.52 (0.04) & 9.12 (0.07) & 2.54 (0.10) & 10.13 (0.06) & 1.34 (0.07)\\
			Regression ($d=20$) & 8.31 (0.06) & 3.42 (0.04) & 8.21 (0.04) & 3.27 (0.03) & 9.54 (0.06) & 2.16 (0.07) & 10.49 (0.02) & 1.03 (0.02)\\
			\bottomrule
		\end{tabular}}
	\end{center}
	\caption{Feature importances in the synthetic data tasks. We sum up the feature importances of all the stable variables and all the environmental variables. In this table, \emph{Stable} stands for the feature importance sum of all the stable variables, and \emph{Environmental} stands for the feature importance sum of all the environmental variables. We take the average of the results over 5 runs. The number in the bracket is the standard deviation of the results over 5 runs. As we can see, when $\lambda$ goes up, IRF uses stable variables more and environmental variables less, as we expect.}
	\label{featureimportance}
\end{table*}

\begin{table*}[t]
	\small
	\begin{center}
		\begin{tabular}{cccccc} 
			\toprule
			Dataset & RF &  XGBoost & IRF ($\lambda = 1$)& IRF ($\lambda = 5$)& IRF ($\lambda = 10$) \\
			\midrule
			Financial & 53.07 (1.06)	& 53.05 (1.27) &55.30 (1.27) & 56.34 (1.42) & \textbf{56.84 (1.47)}  \\
			Technical  &  49.07 (0.29) & 49.10 (0.33) & 49.36 (0.45) & 50.07 (0.67) & \textbf{50.65 (0.75)} \\
			Asset Pricing &51.45 (0.38) & 51.39 (0.26) & \textbf{51.96 (0.39)} & 51.91 (0.41) & 51.85 (0.42) \\
			\bottomrule
		\end{tabular}
	\end{center}
	\caption{Accuracy results of real-data classification tasks in Scenario 1.  Standard deviations are in the brackets.}
	\label{class1}
\end{table*}

\begin{table}[t]
	\small
	\begin{center}
		\begin{tabular}{ccccc} 
			\toprule
			Dataset & RF &  XGBoost & IRF \\
			\midrule
			Financial & 54.42 (1.19)	& 52.46 (1.32) & \textbf{56.20 (1.44)}  \\
			Technical & 48.89 (0.71) & 49.05 (0.72) & \textbf{49.18 (1.14)} \\
			Asset Pricing & 52.50 (0.50) &  51.95  (0.33) &  \textbf{52.92 (0.47)} \\
			\midrule
			Financial & \textbf{54.50 (1.33)} & 52.46 (1.32) & 54.27 (1.22) \\
			Technical & 48.94 (1.79) & 49.05 (1.69) & \textbf{49.57 (2.06)}\\
			Asset Pricing & 52.64 (0.49) & 51.95 (0.33) & \textbf{52.79 (0.44)}\\
			\bottomrule
		\end{tabular}
	\end{center}
	\caption{Accuracy results of real-data classification tasks in Scenario 2 and 3. S2 is the upper part and S3 is the lower part. Standard deviations are in the brackets.}
	\label{class2}
\end{table}

Next, we test the proposed Invariant Random Forest (IRF) on several datasets with different OOD settings. We compare our method with Random Forest (RF) and XGBoost, which are two most popular ensemble tree models that are even being used in industries very often. For synthetic data experiments, we also include the results of linear regression (LR) and two popular OOD generalization methods on neural networks, IRM and REx, for reference. Other methods are not included for comparisons because there are few works considering the OOD generalization of tree models, as far as we know.

If not stated otherwise, for results of classification tasks, the first number in the table represents the average accuracy in percentage, and the number in the bracket is the standard deviation in multiple runs. For results of regression tasks, the first number represents the average Mean Square Error (MSE), and the number in the bracket is the standard deviation in multiple runs. For regression tasks, we standardize the MSE results of RF to 1, and the MSE of every other method is shown using the ratio compared to RF. That is to say, a number in the table of a regression task less than 1 means better performance than RF.

\subsection{Hyperparameter Settings}
\label{sec:hyperparameters}

If there is no validation set, the maximum depths of RF, IRF, and XGBoost are fixed to 10 in classification tasks, and 20 in regression tasks. For each task, we run IRF with $\lambda = 1,5,10$ respectively. If there is a validation set, RF chooses the best maximum depth from $\{ 5,10,15 \}$ (for classification tasks) or $\{10,15,20\}$ (for regression tasks). IRF uses the same maximum depth as RF and chooses the best $\lambda$ from $\{0,1,5,10\}$. XGBoost also chooses the best maximum depth from $\{ 5,10,15 \}$ (for classification tasks) or $\{10,15,20\}$ (for regression tasks). All these hyperparameters are chosen using the cross-entropy loss (for classification tasks) or MSE (for regression tasks) on the validation set.

As for the number of trees, in the cases of Scenario 2 and Scenario 3 in real data regression tasks, the number of ensemble trees is fixed to 10 for RF and IRF. In all the other cases, the number of ensemble trees is fixed to 50 for RF and IRF. For XGBoost, the number of ensemble trees is fixed to 100 in all the experiments without change.

For most of the datasets we use in the experiments, there is no preprocessing. The only exception is the Asset Pricing dataset. As we take rolling windows to train and forecast, and some features are N/A during some time periods, we remove the features which have N/A values on any row in the selected window. Therefore, the feature dimension in training may be less than the feature dimension of the  original dataset, and we may use different features as inputs of models in different  windows.

\subsection{Synthetic Data Experiments}

We design a classification dataset and a regression dataset to validate our idea and method. Both of the datasets allow some difference between the training distribution and the testing distribution. We report the results of IRF when the hyper-parameter $\lambda = 1, 5, 10$.

The data points of the classification task are generated by:
\begin{equation}
\begin{aligned}
    Y & = \text{Bern}(0.5), \\
    X_1 & = |Y \cdot 1_d - C_1| + N_1,\quad C_1 \sim \text{Bern}_d(0.3),\\
    X_2 & = |Y \cdot 1_d - C_2| + N_2, \quad C_2 \sim \text{Bern}_d(U_e),\\
    N_1 & ,~ N_2  \sim \mathcal{N}(0,I_d).
\end{aligned}
\end{equation}
$\text{Bern}_d$ is the $d$-dimensional Bernoulli distribution with each dimension independent. $1_d$ is the $d$-dimensional vector with all entries being 1, and $I_d$ is the $d\times d$ identity matrix.
We set $U_1 = 0.1$, $U_2 = 0.4$, and $U_3 = 0.7$. 
The training data consists of two environments $e=1,2$, and the testing data is the environment $e=3$.
This is an updated version of the example we proposed previously. Both $X_1$ and $X_2$ are $d$-dimensional and are added by a Gaussian noise.
In addition, the data points of the regression task are generated by:
\begin{equation}
	\begin{aligned}
 X_1 & \sim \mathcal{N}(0,I_d), \\
 Y & = 1_d^\top X_1 + N,\quad  N  \sim \mathcal{N}(0,d),\\
 X_2 & =  Y + N_e,\quad N_e \sim \mathcal{N}(0,\sigma_e^2 d).\\
\end{aligned}
\end{equation}
Again the training data is $e=1,2$, and the testing data is the environment $e=3$. We set $\sigma_1 = 0.1$, $\sigma_2 = 2$, and $\sigma_3 = 5$.  In environment $e=1$, $X_2$ can predict $Y$ directly with a relatively low error. But in other environments, $X_2$ has high errors in predicting $Y$. Models that use $X_1$ to predict $Y$ would achieve greater performance in the testing environment, because it is more stable.

The results are shown in Table \ref{synclassification} and \ref{synregression}. In both tasks, IRF shows superior performance compared to the other methods, across different dataset dimensions. This proves that our method can recognize the unstable splits and avoid using these splits during the growing of the trees. Moreover, a large $\lambda$ seems to be a good choice. In real applications, $\lambda$ can be selected with a validation set.
Actually, when $\lambda$ increases, our method uses less of $X_2$ and more of $X_1$ to predict, as illustrated as follows.

In the two synthetic data experiments, except for using the accuracy or MSE on the testing set as evaluation metrics, we can directly check the frequency of each variable being the splitting variable in the tree growth. We want our model to use stable variables more and use environmental variables less. To measure the frequency of a variable used as the splitting variable, we define a feature importance to each feature/variable. For a single tree, the initial feature importances are all 0. Whenever a variable is used as the splitting variable on node $m$, the feature importance of this variable increases by $\frac{n_m}{n}$, where $n_m$ is the sample size on node $m$ and $n$ is the sample size of the whole training set. The feature importance over a forest is the average feature importance over all its trees. We sum up the feature importances of all stable variables and all environmental variables in  the two synthetic data experiments and present them in Table \ref{featureimportance}. As we can see, when $\lambda$ goes up, IRF uses stable variables more and environmental variables less, as we expect.

\subsection{Settings of Real Data Experiments}

\begin{table*}[t]
	\small
	\begin{center}
		\begin{tabular}{cccccc} 
			\toprule
			Scenario & Dataset & Environment 1 & Environment 2 & Environment 3 & Environment 4 \\
			\midrule
			\multirow{16}{*}{S1} & Energy1-1 & hour $ \in [10,18)$ & hour $ \in [6,10) \cup [18,22)$ & hour $ \in [0,6) \cup [22,24)$ \\
			& Energy1-2 & hour $\in [4,12)$ & hour $ \in [12,20)$ & hour $ \in [0,4) \cup [20,24)$ \\
			& Energy1-3 & hour $ \in [0,8)$ & hour $ \in [8,16)$ & hour $ \in [16,24)$ \\
			& Energy1-4 & first $1/3$ of time & middle $1/3$ of time & last $1/3$ of time \\
			& Energy2-1 & hour $ \in [10,18)$ & hour $ \in [6,10) \cup [18,22)$ & hour $\in [0,6) \cup [22,24)$ \\
			& Energy2-2 & hour $ \in [4,12)$ & hour $ \in [12,20)$ & hour $ \in [0,4) \cup [20,24)$ \\
			& Energy2-3 & hour $ \in [0,8)$ & hour $ \in [8,16)$ & hour $ \in [16,24)$ \\
			& Energy2-4 & first $1/3$ of time & middle $1/3$ of time & last $1/3$ of time \\
			& Air1-1 & month $=1,2,3,4$ & month $=5,6,7,8$ & month $=9,10,11,12$ \\
			& Air1-2 & month $=2,3,4,5$ & month $=6,7,8,9$ & month $=10,11,12,1$ \\
			& Air1-3 & month $=3,4,5,6$ & month $=7,8,9,10$ & month $=11,12,1,2$ \\
			& Air1-4 & first $1/3$ of time & middle $1/3$ of time & last $1/3$ of time \\
			& Air2-1 & month $=1,2,3,4$ & month $=5,6,7,8$ & month $=9,10,11,12$ \\
			& Air2-2 & month $=2,3,4,5$ & month $=6,7,8,9$ & month $=10,11,12,1$ \\
			& Air2-3 & month $=3,4,5,6$ & month $=7,8,9,10$ & month $=11,12,1,2$ \\
			& Air2-4 & first $1/3$ of time & middle $1/3$ of time & last $1/3$ of time \\
			
			\midrule
			\multirow{12}{*}{S2 \& S3} & Energy1-1 & hour $\in [0,6)$ & hour $\in [6,12)$ & hour $\in [12,18)$ & hour $\in [18,24)$ \\
			& Energy1-2 &first $1/4$ of time & second $1/4$ of time & third $1/4$ of time & last $1/4$ of time\\
			& Energy2-1 & hour $\in [0,6)$ & hour $\in [6,12)$ & hour $\in [12,18)$ & hour $\in [18,24)$ \\
			& Energy2-2 &first $1/4$ of time & second $1/4$ of time & third $1/4$ of time & last $1/4$ of time\\
			& Air1-1 & month $=1,2,3$ & month $=4,5,6$ & month $=7,8,9$ & month $=10,11,12$ \\
			& Air1-2 & month $=2,3,4$ & month $=5,6,7$ & month $=8,9,10$ & month $=11,12,1$ \\
			& Air1-3 & month $=3,4,5$ & month $=6,7,8$ & month $=9,10,11$ & month $=12,1,2$ \\
			& Air1-4 &first $1/4$ of time & second $1/4$ of time & third $1/4$ of time & last $1/4$ of time\\
			& Air2-1 & month $=1,2,3$ & month $=4,5,6$ & month $=7,8,9$ & month $=10,11,12$ \\
			& Air2-2 & month $=2,3,4$ & month $=5,6,7$ & month $=8,9,10$ & month $=11,12,1$ \\
			& Air2-3 & month $=3,4,5$ & month $=6,7,8$ & month $=9,10,11$ & month $=12,1,2$ \\
			& Air2-4 &first $1/4$ of time & second $1/4$ of time & third $1/4$ of time & last $1/4$ of time\\
			\bottomrule
		\end{tabular}
	\end{center}
	\caption{Splitting strategies of regression datasets. Each dataset is split into three or four environments by the strategies in the table. The training set, validation set, and testing set are constructed using these environments.}
	\label{regressionsplit}
\end{table*}

\begin{table*}[t]
\small
\begin{center}
\begin{tabular}{cccccc} 
\toprule
 Dataset & RF &  XGBoost & IRF ($\lambda$ = 1) & IRF ($\lambda$ = 5) & IRF ($\lambda$ = 10)\\
\midrule
Energy1-1  &  \textbf{1.000} & 1.266 (0.153) & 1.009 (0.007) & 1.029 (0.016) & 1.027 (0.024) \\
Energy1-2 & 1.000 & 1.201 (0.102) & 0.986 (0.027) & 0.961 (0.086) &  \textbf{0.958 (0.109) }\\
Energy1-3 &  \textbf{1.000 }& 1.219 (0.106) & 1.018 (0.012) & 1.032 (0.004) & 1.045 (0.023)\\
Energy1-4 & 1.000 & 1.287 (0.125) & 0.933 (0.084) & 0.896 (0.129) &  \textbf{0.892 (0.129)}\\
Energy2-1 & 1.000 & 1.186 (0.095) & 0.891 (0.131) & 0.862 (0.206) &  \textbf{0.854 (0.218)}\\
Energy2-2 & 1.000 & 1.088 (0.022) & 0.997 (0.017) & 0.976 (0.083) &  \textbf{0.974 (0.094)}\\
Energy2-3 & 1.000 & 1.144 (0.103) &  \textbf{0.973 (0.015)} & 1.019 (0.075) & 1.054 (0.091) \\
Energy2-4 & 1.000 & 1.010 (0.063) & 0.988 (0.019) & 1.003 (0.007) &  \textbf{0.983 (0.039)} \\
Air1-1 & 1.000 & 1.130 (0.015) & 0.878 (0.122) &  \textbf{0.850 (0.200)} & 0.865 (0.222)\\
Air1-2 & 1.000 & 1.088 (0.043) & 0.969 (0.016) & 0.911 (0.025) &  \textbf{0.886 (0.063)} \\
Air1-3 & 1.000 & 1.107 (0.076) & 0.983 (0.027) & 0.944 (0.085) &  \textbf{0.923 (0.084)} \\
Air1-4 & 1.000 & 1.243 (0.061) &  \textbf{0.901 (0.148)} & 0.974 (0.228) & 1.006 (0.246)\\
Air2-1 & 1.000 & 1.214 (0.084) & 0.987 (0.003) &  \textbf{0.976 (0.024)} & 0.995 (0.043) \\
Air2-2 & 1.000 & 1.260 (0.122) &  \textbf{0.981 (0.033)} & 1.013 (0.064) & 1.092 (0.068) \\
Air2-3 & 1.000 & 1.242 (0.144) & 0.973 (0.020) &  \textbf{0.965 (0.035)} & 0.984 (0.026) \\
Air2-4 & 1.000 &  \textbf{0.858 (0.020)} & 1.037 (0.009) & 1.076 (0.012) & 1.057 (0.010)\\ 
\bottomrule
\end{tabular}
\end{center}
\caption{MSE results of real-data regression tasks in Scenario 1. Standard deviations are in
	the brackets.}
\label{regression1}
\end{table*}

\begin{table}[t]
	\small
	\begin{center}
		\begin{tabular}{cccc} 
			\toprule
			Dataset & RF &  XGBoost & IRF \\
			\midrule
			Energy1-1 & 1.000 & 1.064 (0.113) & \textbf{0.989 (0.047)} \\
			Energy1-2 & 1.000 & 1.141 (0.231) &  \textbf{0.925 (0.116}) \\
			Energy2-1 &  \textbf{1.000} & 1.027 (0.107) & 1.006 (0.050) \\
			Energy2-2 &  \textbf{1.000} & 1.059 (0.103) & 1.037 (0.121) \\
			Air1-1 & 1.000 & 1.052 (0.099) &  \textbf{0.916 (0.106)} \\ 
			Air1-2 & 1.000 & 1.112 (0.097) &  \textbf{0.979 (0.064)} \\
			Air1-3 & 1.000 & 1.173 (0.191) &  \textbf{0.940 (0.141)}\\
			Air1-4 & \textbf{1.000} & 1.099 (0.042) & 1.006 (0.013) \\
			Air2-1 & 1.000 & 1.187 (0.139) &  \textbf{0.970 (0.078)}\\
			Air2-2 & 1.000 & 1.191 (0.117) &  \textbf{0.982 (0.062)}\\
			Air2-3 & 1.000 & 1.236 (0.109) &  \textbf{0.986 (0.042)}\\
			Air2-4 & 1.000 & 1.089 (0.084) & \textbf{0.995 (0.031)}\\
			\midrule
			Energy1-1  & 1.000 & 1.050 (0.086) & \textbf{0.977 (0.056)}  \\
			Energy1-2 & 1.000 & 1.187 (0.200) & \textbf{0.979 (0.142)}  \\
			Energy2-1 & \textbf{1.000}  & 1.008 (0.097) & 1.038 (0.143) \\
			Energy2-2 & \textbf{1.000}  & 1.068 (0.099) & 1.062 (0.174) \\
			Air1-1 & 1.000 & 1.053 (0.076) & \textbf{0.845 (0.230)} \\
			Air1-2 & 1.000 & 1.126 (0.076) & \textbf{0.956 (0.101)} \\
			Air1-3 & 1.000 & 1.151 (0.202) & \textbf{0.988 (0.072)} \\
			Air1-4 & 1.000 & 1.110 (0.046) & \textbf{0.993 (0.010)} \\
			Air2-1 & 1.000 & 1.173 (0.115) & \textbf{0.926 (0.180)} \\
			Air2-2 & \textbf{1.000}  & 1.205 (0.126) & 1.005 (0.070)\\
			Air2-3 & \textbf{1.000}  & 1.223 (0.108) & 1.005 (0.055)\\
			Air2-4 & 1.000 & 1.091 (0.076) & \textbf{0.991 (0.031)} \\
			\bottomrule
		\end{tabular}
	\end{center}
	\caption{MSE results of real-data regression tasks in Scenario 2 and 3. Standard deviations are in
		the brackets.  S2 is the upper part and S3 is the lower part.}
	\label{regression2}
\end{table}

In the next, we test our method on real datasets, to see the practicability of IRF on real prediction applications.
To run the experiments, we first split a dataset into various environments according to some basic variables that contain meta information (year, month, timestamps, etc.). 
When using such a variable and partitioning it to obtain  environments, this variable will not be used as the inputs for models. To test the OOD performance of models, we use one environment as the testing data. During training, we have no access to this environment at all. For training data, we consider three different training scenarios based on whether the environment information is given:

(S1) Only the pooled data of all environments excluding the testing one is provided as training data, i.e.,  we do not know any environmental information on training data.

(S2) The pooled data of most environments is provided, but we have a validation set from a brand new environment, which can be used for hyper-parameter tuning. 

(S3) Training data with full environmental information is provided. That is to say, we take multiple groups of data as inputs of models, and each group is a single environment. 

The last scenario is more popular in research works, but the first two scenarios are realistic too because sometimes we do not know any meta information.
In such cases, we manually split the training data and obtain hand-made environments with a recent simple but effective method of doing so, called Decorr \cite{liao2022decorr,ye2023coping,tong2023quantitatively}. It finds subsets of training data that exhibit low correlations among features, for reducing spurious correlations in the data.
While other methods of the same purpose mainly focus on DNNs, Decorr has no restrictions.

The ways of obtaining initial environment partitions and the settings of training and testing sets are different for classification and regression. We introduce them separately in the following subsections. 
The average results of IRF from multiple runs of training and testing are reported and are compared to RF and XGBoost.
In Scenario 1, there is no hyper-parameter tuning and we set $\lambda = 1, 5, 10$, respectively.
For the validation procedure in Scenario 2 and 3, we first choose the proper maximum tree depth for RF on validation set. Then, we use the same maximum depth in IRF and use the validation set to choose the best $\lambda$ for IRF.  For details regarding the validation procedure and the choices of some other hyper-parameters, please go to Section \ref{sec:hyperparameters}.

\subsection{Real-Data Classification Tasks}

Three datasets are included in this study here. All of them have heterogeneous distributions or prediction patterns during different periods of time. So, we initially obtain the environments by splitting the datetime.
The introduction of each dataset is in the following.
The results are shown in Table \ref{class1} and Table \ref{class2}, where IRF performs the best overall than others.

\paragraph{Financial}
Financial indicator dataset\footnote{https://www.kaggle.com/datasets/cnic92/200-financial-indicators-of-us-stocks-20142018} is a yearly stock return dataset. The features are the financial indicators of each stock, and the label we need to predict is the price going up or down in the next whole year. The span of this dataset is 5 years, and we treat each single year as an environment. In Scenario 1, we use any 3 environments as the training set, and the other two as the testing set. In Scenario 2 and 3, we use any 3 environments as the training set, 1 as the validation set, and 1 as the testing set.

\paragraph{Technical}
Technical indicator dataset\footnote{https://www.kaggle.com/datasets/nikhilkohli/us-stock-market-data-60-extracted-features} contains market data and technical indicators of 10 U.S. stocks from 2005 to 2020. The model is expected to predict if tomorrow's close price of a stock is higher than today's. We split this dataset into several environments based on  different time periods. In Scenario 1, data of the first 60\% of the time is used as the training set, and the rest as the testing set. In Scenario 2 and 3, we use the first 60\% of the data as the training set (split into two environments as the first half and the second half, if needed), then the subsequent 20\% as the validation set, and the last 20 \% as the testing set.

\paragraph{Asset Pricing}
Asset Pricing \cite{gu2020empirical,gu2021autoencoder} is a monthly stock return dataset with different types of financial and technical indicators for thousands of stocks during the period from 1970 to 2021. Models are expected to predict if one stock has a higher monthly return than at least a half of all stocks in the market at the same time. 
In Scenario 1, we adopt 9-month rolling windows for training and testing. The first six months is the training set, and the next three months is the testing set. In Scenario 2 and 3, we adopt one-year rolling windows. The first six months is the training set again (split into the first three months and the last three as two environments, if needed), the next three months is the validation set, and the last three is the testing set.

\subsection{Real-Data Regression Tasks}

The training and testing settings of regression tasks all follow a same strategy.
For the first scenario, we obtain three environments using the variable containing meta information first. We further use any two environments as training set, and the other one as testing set (3 runs for each dataset). For the second and third scenarios, we obtain four environments initially and use any two as training set, any other one as validation set, and the remaining one as testing set (12 runs for each dataset).
In this way, the training set and testing set will have heterogeneous distributions, hence the performance on the testing set can reflect the OOD generalization ability of the model.
We introduce the four datasets used as follow.

\paragraph{Energy1 \& Energy2}
Energy datasets\footnote{https://github.com/LuisM78/Appliances-energy-prediction-data/blob/master/energydata\_complete.csv, https://www.kaggle.com/datasets/gmkeshav/tetuan-city-power-consumption} provide some information like temperature, humidity, and windspeed in a city or a room. We want to predict the energy consumption of that location. Since time has a heterogeneous influence on the matter of energy consumption (e.g., a big difference between day and night), in both datasets, we use the variable \emph{hour} to split the data and obtain environments. There are four different splitting strategies considered for Scenario 1 and two strategies for Scenario 2 and 3, resulting in more \emph{derived datasets} named as Energy1-$i$ (or Energy2-$i$). The details of the splitting strategies are in Table \ref{regressionsplit}.

\paragraph{Air1 \& Air2}
Air datasets\footnote{https://github.com/sunilmallya/timeseries/blob/master/data/ \\ PRSA\_data\_2010.1.1-2014.12.31.csv, } contain various types of air quality data, and models need to predict the PM2.5 index based on the air quality data given. In these two datasets, we use the variable \emph{month}  to split and obtain environments, because in different seasons, the behaviors of air pollution may be different. We again have different splitting strategies considered (four for Scenario 1, 2, and 3), resulting in more \emph{derived datasets} named as Air1-$i$ (or Air2-$i$), as shown in Table \ref{regressionsplit}.

\paragraph{Results} 
The results are shown in Table \ref{regression1} and Table \ref{regression2}, where IRF wins the best performance overall.
On many datasets, the performance improvements of IRF over the other two are much significant, with a maximum percentage decrease close to 15\%. In many cases, it ranges from 1\% to 9\%.
Another finding is that XGBoost always performs worse than RF in all OOD generalization tasks in this paper, indicating that bagging may be better than boosting in OOD settings.

\section{Conclusion}
Through the discovery of the invariant when using stable variables as the splitting variable of a tree, we construct a method for tree-based models to reduce the use of environmental variables by enforcing penalties when splitting.
The proposed IDT and IRF are well-motivated, easy to interpret, and new to the area.
Experiments have shown that our method achieves superior performance under the OOD generalization settings. 

Some topics are still worth discussing on our method. For example, 
the selection of hyper-parameter $\lambda$ when there are no validation sets, e.g., in Scenario 1. 
It may not be a good choice if the validation set is \emph{i.i.d.} extracted from the training set.
Since many existing works assume the existence of meta information,
we will leave the undiscussed topics as our future work. And other methodologies for OOD generalization of tree models are worth being explored.

%
%

\bibliography{aaai24}

\begin{thebibliography}{25}
\providecommand{\natexlab}[1]{#1}

\bibitem[{Arjovsky et~al.(2019)Arjovsky, Bottou, Gulrajani, and
  Lopez-Paz}]{arjovsky2019invariant}
Arjovsky, M.; Bottou, L.; Gulrajani, I.; and Lopez-Paz, D. 2019.
\newblock Invariant risk minimization.
\newblock \emph{arXiv preprint arXiv:1907.02893}.

\bibitem[{Breiman(2001)}]{breiman2001random}
Breiman, L. 2001.
\newblock Random forests.
\newblock \emph{Machine learning}, 45: 5--32.

\bibitem[{Chen et~al.(2015)Chen, He, Benesty, Khotilovich, Tang, Cho, Chen,
  Mitchell, Cano, Zhou et~al.}]{chen2015xgboost}
Chen, T.; He, T.; Benesty, M.; Khotilovich, V.; Tang, Y.; Cho, H.; Chen, K.;
  Mitchell, R.; Cano, I.; Zhou, T.; et~al. 2015.
\newblock Xgboost: extreme gradient boosting.
\newblock \emph{R package version 0.4-2}, 1(4): 1--4.

\bibitem[{Cui and Athey(2022)}]{cui2022stable}
Cui, P.; and Athey, S. 2022.
\newblock Stable learning establishes some common ground between causal
  inference and machine learning.
\newblock \emph{Nature Machine Intelligence}, 4(2): 110--115.

\bibitem[{Friedman(2001)}]{friedman2001greedy}
Friedman, J.~H. 2001.
\newblock Greedy function approximation: a gradient boosting machine.
\newblock \emph{Annals of statistics}, 1189--1232.

\bibitem[{Geirhos et~al.(2020)Geirhos, Jacobsen, Michaelis, Zemel, Brendel,
  Bethge, and Wichmann}]{geirhos2020shortcut}
Geirhos, R.; Jacobsen, J.-H.; Michaelis, C.; Zemel, R.; Brendel, W.; Bethge,
  M.; and Wichmann, F.~A. 2020.
\newblock Shortcut learning in deep neural networks.
\newblock \emph{Nature Machine Intelligence}, 2(11): 665--673.

\bibitem[{Gordon et~al.(1984)Gordon, Breiman, Friedman, Olshen, and
  Stone}]{gordon1984classification}
Gordon, A.; Breiman, L.; Friedman, J.; Olshen, R.; and Stone, C.~J. 1984.
\newblock Classification and Regression Trees.
\newblock \emph{Biometrics}, 40(3): 874.

\bibitem[{Gu, Kelly, and Xiu(2020)}]{gu2020empirical}
Gu, S.; Kelly, B.; and Xiu, D. 2020.
\newblock Empirical asset pricing via machine learning.
\newblock \emph{The Review of Financial Studies}, 33(5): 2223--2273.

\bibitem[{Gu, Kelly, and Xiu(2021)}]{gu2021autoencoder}
Gu, S.; Kelly, B.; and Xiu, D. 2021.
\newblock Autoencoder asset pricing models.
\newblock \emph{Journal of Econometrics}, 222(1): 429--450.

\bibitem[{Krueger et~al.(2021)Krueger, Caballero, Jacobsen, Zhang, Binas,
  Zhang, Le~Priol, and Courville}]{krueger2021out}
Krueger, D.; Caballero, E.; Jacobsen, J.-H.; Zhang, A.; Binas, J.; Zhang, D.;
  Le~Priol, R.; and Courville, A. 2021.
\newblock Out-of-distribution generalization via risk extrapolation (rex).
\newblock In \emph{International Conference on Machine Learning}, 5815--5826.
  PMLR.

\bibitem[{Kuang et~al.(2018)Kuang, Cui, Athey, Xiong, and Li}]{kuang2018stable}
Kuang, K.; Cui, P.; Athey, S.; Xiong, R.; and Li, B. 2018.
\newblock Stable prediction across unknown environments.
\newblock In \emph{proceedings of the 24th ACM SIGKDD international conference
  on knowledge discovery \& data mining}, 1617--1626.

\bibitem[{Kuang et~al.(2020)Kuang, Xiong, Cui, Athey, and Li}]{kuang2020stable}
Kuang, K.; Xiong, R.; Cui, P.; Athey, S.; and Li, B. 2020.
\newblock Stable prediction with model misspecification and agnostic
  distribution shift.
\newblock In \emph{Proceedings of the AAAI Conference on Artificial
  Intelligence}, volume~34, 4485--4492.

\bibitem[{Liao, Wu, and Yan(2022)}]{liao2022decorr}
Liao, Y.; Wu, Q.; and Yan, X. 2022.
\newblock Decorr: Environment Partitioning for Invariant Learning and OOD
  Generalization.
\newblock \emph{arXiv preprint arXiv:2211.10054}.

\bibitem[{Liu et~al.(2021)Liu, Shen, Cui, Zhou, Kuang, Li, and
  Lin}]{liu2021stable}
Liu, J.; Shen, Z.; Cui, P.; Zhou, L.; Kuang, K.; Li, B.; and Lin, Y. 2021.
\newblock Stable adversarial learning under distributional shifts.
\newblock In \emph{Proceedings of the AAAI Conference on Artificial
  Intelligence}, volume~35, 8662--8670.

\bibitem[{Moneda and Mau{\'a}(2022)}]{moneda2022time}
Moneda, L.; and Mau{\'a}, D. 2022.
\newblock Time Robust Trees: Using Temporal Invariance to Improve
  Generalization.
\newblock In \emph{Brazilian Conference on Intelligent Systems}, 385--397.
  Springer.

\bibitem[{Quinlan(2014)}]{quinlan2014c4}
Quinlan, J.~R. 2014.
\newblock \emph{C4. 5: programs for machine learning}.
\newblock Elsevier.

\bibitem[{Rosenfeld, Ravikumar, and Risteski(2020)}]{rosenfeld2020risks}
Rosenfeld, E.; Ravikumar, P.~K.; and Risteski, A. 2020.
\newblock The Risks of Invariant Risk Minimization.
\newblock In \emph{International Conference on Learning Representations}.

\bibitem[{Shen et~al.(2020{\natexlab{a}})Shen, Cui, Liu, Zhang, Li, and
  Chen}]{shen2020stable}
Shen, Z.; Cui, P.; Liu, J.; Zhang, T.; Li, B.; and Chen, Z. 2020{\natexlab{a}}.
\newblock Stable learning via differentiated variable decorrelation.
\newblock In \emph{Proceedings of the 26th acm sigkdd international conference
  on knowledge discovery \& data mining}, 2185--2193.

\bibitem[{Shen et~al.(2020{\natexlab{b}})Shen, Cui, Zhang, and
  Kunag}]{shen2020stable_1}
Shen, Z.; Cui, P.; Zhang, T.; and Kunag, K. 2020{\natexlab{b}}.
\newblock Stable learning via sample reweighting.
\newblock In \emph{Proceedings of the AAAI Conference on Artificial
  Intelligence}, volume~34, 5692--5699.

\bibitem[{Shen et~al.(2021)Shen, Liu, He, Zhang, Xu, Yu, and
  Cui}]{shen2021towards}
Shen, Z.; Liu, J.; He, Y.; Zhang, X.; Xu, R.; Yu, H.; and Cui, P. 2021.
\newblock Towards out-of-distribution generalization: A survey.
\newblock \emph{arXiv preprint arXiv:2108.13624}.

\bibitem[{Tong et~al.(2023)Tong, Yuan, Zhang, Zhu, Zhang, Wu, and
  Kuang}]{tong2023quantitatively}
Tong, Y.; Yuan, J.; Zhang, M.; Zhu, D.; Zhang, K.; Wu, F.; and Kuang, K. 2023.
\newblock Quantitatively Measuring and Contrastively Exploring Heterogeneity
  for Domain Generalization.
\newblock \emph{arXiv preprint arXiv:2305.15889}.

\bibitem[{Ye et~al.(2023)Ye, Yu, Hou, Wang, and You}]{ye2023coping}
Ye, S.; Yu, S.; Hou, W.; Wang, Y.; and You, X. 2023.
\newblock Coping with Change: Learning Invariant and Minimum Sufficient
  Representations for Fine-Grained Visual Categorization.
\newblock \emph{arXiv preprint arXiv:2306.04893}.

\bibitem[{Yu et~al.(2023)Yu, Cui, He, Shen, Lin, Xu, and Zhang}]{yu2023stable}
Yu, H.; Cui, P.; He, Y.; Shen, Z.; Lin, Y.; Xu, R.; and Zhang, X. 2023.
\newblock Stable Learning via Sparse Variable Independence.
\newblock In \emph{Proceedings of the AAAI Conference on Artificial
  Intelligence}, volume~37, 10998--11006.

\bibitem[{Zhang et~al.(2021)Zhang, Cui, Xu, Zhou, He, and Shen}]{zhang2021deep}
Zhang, X.; Cui, P.; Xu, R.; Zhou, L.; He, Y.; and Shen, Z. 2021.
\newblock Deep stable learning for out-of-distribution generalization.
\newblock In \emph{Proceedings of the IEEE/CVF Conference on Computer Vision
  and Pattern Recognition}, 5372--5382.

\bibitem[{Zhou et~al.(2022)Zhou, Liu, Qiao, Xiang, and Loy}]{zhou2021domain}
Zhou, K.; Liu, Z.; Qiao, Y.; Xiang, T.; and Loy, C.~C. 2022.
\newblock Domain Generalization: A Survey.
\newblock \emph{IEEE Transactions on Pattern Analysis and Machine
  Intelligence}.

\end{thebibliography}

%
%
%

%
%
%

\end{document}